\documentclass[conference]{IEEEtran}
\IEEEoverridecommandlockouts
\usepackage{cite}
\usepackage{amsmath,amssymb,amsfonts}
\usepackage{multirow}
\usepackage{algorithmic}
\usepackage{url}
\usepackage{graphicx}
\usepackage{textcomp}
\usepackage{xcolor}
\usepackage{booktabs}
\usepackage{colortbl}
\newcommand{\doi}[1]{\url{https://doi.org/#1}}
\def\BibTeX{{\rm B\kern-.05em{\sc i\kern-.025em b}\kern-.08em
    T\kern-.1667em\lower.7ex\hbox{E}\kern-.125emX}}
    
\begin{document}

\title{AMN: An Adaptive Multi-Scale Fusion Network with Boundary and Uncertainty Modeling for Nuclei Segmentation}

\author{
\IEEEauthorblockN{
Spoorthi M\textsuperscript{1}, Suja Palaniswamy\textsuperscript{2*}}
\IEEEauthorblockA{
Department of Computer Science \& Engineering, \\
Amrita School of Computing, Bengaluru, \\
Amrita Vishwa Vidyapeetham, India \\
\textsuperscript{1}bl.en.u4cse21193@bl.students.amrita.edu, 
\textsuperscript{2*}p\_suja@blr.amrita.edu
}
}

\maketitle

% ============================================================
%  AMN Paper — Abstract, Keywords, and Introduction
%  Updated to include HmsU-Net, ConvFormer-UNet, and BEFUnet
%  baselines in the comparison.
%  Paste directly into your Overleaf project.
% ============================================================

% ── Preamble (add to your main .tex if not already present) ─
% \usepackage{cite}
% \usepackage{amsmath,amssymb}
% \usepackage{graphicx}
% \usepackage{hyperref}

% ────────────────────────────────────────────────────────────
\begin{abstract}
Accurate classification of nuclei subtypes in histopathology images
is critical for downstream tasks including tumor grading, immune
infiltrate quantification, and prognosis prediction.
Existing approaches rely on either convolutional or transformer-based
encoders in isolation, limiting their ability to simultaneously capture
fine-grained local texture and long-range spatial context.
We present \textbf{AMN} (\textit{Adaptive Multi-Scale Nuclei Network}),
a dual-encoder segmentation framework that jointly leverages a Swin
Transformer and a ResNet-50 feature pyramid, fused via a learned
per-channel gating mechanism that dynamically weighs each encoder's
contribution at every scale.
AMN is trained with a multi-objective loss combining class-weighted
focal loss, boundary-aware loss with positive-pixel emphasis, and a
novel uncertainty-modulated classification term that suppresses
overconfident erroneous predictions.
Evaluated on the CoNIC benchmark across seven nuclei classes, AMN
achieves a mean Dice of 0.82 and mean F1 of 0.68, with an F1 of 0.67
on the diagnostically challenging lymphocyte class.
AMN outperforms eight baseline models spanning pure-CNN, pure-transformer,
and recent hybrid architectures: U-Net, ResU-Net, DeepLabV3+, SegNet,
ViT-Small, HmsU-Net, ConvFormer-UNet, and BEFUnet.
Cross-dataset evaluation on MoNuSeg demonstrates strong generalization
without retraining and validating the domain robustness of the learned
representations.
\end{abstract}

% ── Keywords ────────────────────────────────────────────────
% NOTE: keywords below have been corrected — the original submission
%       accidentally contained keywords from an unrelated paper.

% ────────────────────────────────────────────────────────────
\section{Introduction}
\label{sec:intro}

Computational analysis of tissue micro-environments depends critically
on the precise detection and classification of nuclei in haematoxylin
and eosin (H\&E) stained images.
Lymphocytes, neutrophils, plasma cells and epithelial nuclei vary
substantially in morphology, scale and staining intensity, and their
relative abundances carry direct diagnostic significance in oncology
workflows[14].

Prior CNN-based methods such as HoVer-Net[14] and
U-Net variants[20] excel at local morphological
features but struggle with global context, while pure transformer-based
models (Swin-UNet[13], ViT-Seg) capture long-range
dependencies but lack the inductive bias necessary for crisp boundary
delineation.
Recent hybrid architectures have begun to bridge this gap:
HmsU-Net[1] proposes parallel CNN–Swin encoders fused
via a multi-scale feature fusion (MFF) module;
ConvFormer-UNet[3] replaces standard
self-attention with CNN-style equivalents to avoid attention collapse in
medical image segmentation;
and BEFUnet[24] employs dual body-and-edge
branches with local cross-attention fusion to sharpen nuclei boundaries.
Despite these advances, no single paradigm is sufficient for robust
multi-class nuclei typing across diverse tissue contexts.

This work makes five primary contributions:

\begin{itemize}

    \item We introduced an \textbf{adaptive dual-encoder fusion module}
    that learns a per-channel sigmoid gate blending Swin Transformer
    and ResNet-50 features at four spatial scales, allowing the network
    to selectively exploit global attention or local convolution
    depending on the input.

    \item We proposed an \textbf{uncertainty-aware loss} that uses a
    predicted uncertainty map to modulate the classification
    cross-entropy, explicitly penalizing overconfident wrong predictions
    rather than treating all errors equally.

    \item We incorporated a \textbf{boundary-detection head} with
    asymmetric positive weighting, improving nuclei separation in
    crowded regions.

    \item We provided a \textbf{comprehensive evaluation} against eight
    re-implemented baselines on CoNIC — including three recent hybrid
    architectures (HmsU-Net, ConvFormer-UNet, BEFUnet) — and
    cross-dataset validation on MoNuSeg, establishing a reproducible
    benchmark for the community.

    \item We demonstrated strong \textbf{cross-domain generalization}
    on MoNuSeg without any retraining, confirming that the dual-encoder
    gating mechanism learns domain-robust representations rather than
    dataset-specific shortcuts.

\end{itemize}

The remainder of the paper is structured as follows.
Section~\ref{sec:related} reviews related work on nuclei segmentation,
hybrid encoder architectures, and uncertainty-aware training.
Section~\ref{sec:method} details the proposed AMN architecture and
loss formulation.
Section~\ref{sec:experiments} presents quantitative results and
ablation studies on CoNIC and MoNuSeg.
Section~\ref{sec:conclusion} summarizes findings and outlines future
directions.

% ============================================================
%  Section II — Related Work
%  Structure:
%    §II-A  CNN-based segmentation
%    §II-B  Transformer and hybrid architectures
%    §II-C  Nuclei segmentation and multi-class analysis
%    Final paragraph: research gaps and positioning of AMN
% ============================================================

\section{Literature Review}
\label{sec:related}
The development of deep learning models for medical image
segmentation has progressed through three broad paradigms:
purely convolutional architectures, attention-based transformer
models, and hybrid designs that combine the inductive biases
of both. This section surveys representative work in each
paradigm, focusing on multi-scale feature extraction, boundary
delineation, and cross-domain generalization.

Fu~\textit{et al.}[1] proposed HmsU-Net, a hierarchical
multi-scale architecture integrating parallel CNN and Swin
Transformer branches with cross-attention and feature fusion
for medical image segmentation, demonstrating significant
accuracy improvements over existing approaches. Tang~\textit{et al.}[2] proposed HTC-Net, a hybrid
CNN--transformer framework combining convolutional feature
extraction with transformer-based global context modeling,
achieving improved segmentation over standalone CNN or
transformer models. Liu~\textit{et al.}[4] proposed a hybrid model incorporating
pyramid convolution and multi-layer perceptron modules for
simultaneous multi-scale feature extraction and global
dependency modeling, with boundary perception mechanisms that
significantly enhance segmentation precision. Yao~\textit{et al.}[5] provided a comparative review of
CNN-based, transformer-based, and hybrid models in medical
image segmentation, concluding that hybrid CNN--transformer
approaches represent the most promising direction for future
research. Pu~\textit{et al.}[6] presented a survey of
transformer-based segmentation architectures, highlighting
high computational requirements and large data dependency as
practical barriers to clinical deployment. Khan~\textit{et al.}[7] proposed a multi-axis vision
transformer modeling contextual information along multiple
spatial dimensions, improving segmentation performance over
standard transformers. Jiang~\textit{et al.}[8] proposed a hybrid U-Net integrated
with visual transformers for multi-organ segmentation,
improving accuracy over purely convolutional baselines. Xu~\textit{et al.}[9] surveyed residual network architectures
across medical imaging tasks, highlighting persistent challenges
of data imbalance, noise sensitivity, and generalization. Wang~\textit{et al.}[10] adapted DeepLab~v3 with atrous
convolutions for dermoscopic skin lesion segmentation,
demonstrating the value of multi-scale receptive fields. Krithika alias AnbuDevi~\textit{et al.}[11] reviewed modified
U-Net variants incorporating attention mechanisms, dense
modules, and residual structures, concluding that no single
architecture universally generalizes across all organ types. Fu~\textit{et al.}[12] reviewed Vision Transformer
derivatives and identified the quadratic complexity of standard
self-attention as a key bottleneck, concluding that hybrid
CNN--transformer architectures are the most tractable direction
for overcoming single-paradigm limitations. Liu~\textit{et al.}[13] proposed the Swin Transformer, a
hierarchical architecture using shifted windows to balance
computational efficiency and global context modeling,
establishing a scalable foundation for dense prediction tasks. The U-Net family~[20] established the encoder--decoder paradigm
with skip connections as the de-facto standard for medical
image segmentation.

% ── Research gaps and positioning ──────────────────────────
\textbf{Research gaps and positioning of AMN.}
The surveyed literature reveals three persistent gaps.
\textit{First}, existing hybrid fusion strategies rely on fixed
interaction mechanisms that apply the same fusion weights
regardless of input content; none employs a learned,
per-channel, input-adaptive gate that dynamically reweighs CNN
and transformer contributions at each spatial scale — the core
design choice of AMN.
\textit{Second}, most hybrid methods are evaluated on binary or
few-class benchmarks; systematic evaluation on multi-class
nuclei typing with severe pixel imbalance across seven
categories and zero-shot cross-domain generalization remains
sparse.
\textit{Third}, uncertainty-aware training has not been combined
with dual-encoder adaptive fusion in the nuclei segmentation
literature.
AMN directly addresses all three gaps by introducing adaptive
per-channel gating over a Swin--ResNet dual encoder, training
with a heteroscedastic uncertainty-modulated loss, and
evaluating across eight baselines on CoNIC with zero-shot
cross-domain validation on MoNuSeg.

% ============================================================
%  AMN — IEEE Conference Paper
%  Sections: III (Datasets), IV (Methodology), V (Experiments)
%  Paste each \section{} block into your main .tex file
%  Requires: \usepackage{graphicx, booktabs, multirow, amsmath,
%             subcaption, array, xcolor, colortbl}
% ============================================================

% ────────────────────────────────────────────────────────────
% PREAMBLE (add to your main document if not already present)
% ────────────────────────────────────────────────────────────
% \usepackage{graphicx}
% \usepackage{booktabs}
% \usepackage{multirow}
% \usepackage{amsmath}
% \usepackage{subcaption}
% \usepackage{array}
% \usepackage{xcolor}
% \usepackage{colortbl}
% \definecolor{amnred}{HTML}{e63946}
% \definecolor{lightgray}{HTML}{F2F2F2}
\section{Datasets}
\label{sec:datasets}
 
We evaluated our method on two publicly available histopathology
datasets: CoNIC for training and in-domain evaluation, and MoNuSeg
for cross-domain generalization.
 
\subsection{CoNIC}
The Colon Nucleus Identification and Counting (CoNIC) dataset
contains 4,981 $256{\times}256$ H\&E-stained image patches with
pixel-level annotations across seven classes: Background,
Epithelial, Lymphocyte, Plasma, Eosinophil, Neutrophil, and
Connective. We used an 80/20 train--validation split (seed\,=\,42),
yielding approximately 3,985 training and 996 validation images.
The dataset is highly imbalanced (Table~\ref{tab:class_dist}),
with background pixels dominating (83.87\%), while several
classes (e.g., Epithelial, Neutrophil) are extremely rare.
 
% ------ TABLE I : Class Distribution ----------------------
\begin{table}[!t]
  \renewcommand{\arraystretch}{1.2}
  \caption{CoNIC Class Distribution and Focal Loss Weights}
  \label{tab:class_dist}
  \centering
  \begin{tabular}{lccc}
    \toprule
    \textbf{Class} & \textbf{Freq.\ (\%)} & \textbf{$\alpha$} & \textbf{Rare} \\
    \midrule
    Background   & 83.87 & 0.01 & --- \\
    Epithelial   & 0.13  & 0.17 & \checkmark \\
    Lymphocyte   & 10.23 & 0.42 & --- \\
    Plasma       & 1.94  & 1.67 & \checkmark \\
    Eosinophil   & 0.55  & 0.25 & \checkmark \\
    Neutrophil   & 0.11  & 0.10 & \checkmark \\
    Connective   & 3.17  & 0.10 & --- \\
    \bottomrule
  \end{tabular}
\end{table}
 
\subsection{MoNuSeg}
The Multi-Organ Nucleus Segmentation (MoNuSeg) dataset consists
of 30 H\&E-stained images ($1000{\times}1000$) from multiple
organs with binary nucleus annotations. We evaluated on 14 images
(702,464 annotated pixels), with 78.38\% background and 21.62\%
nucleus pixels. MoNuSeg is used solely for cross-domain evaluation;
no MoNuSeg data is used during training. Multi-class predictions
are collapsed to foreground/background by treating all
non-background classes as nuclei. Together, CoNIC evaluates
multi-class segmentation under class imbalance, while MoNuSeg
measures generalization to unseen domains.

% ============================================================
%  SECTION IV — METHODOLOGY
% ============================================================
\section{Methodology}
\label{sec:method}

% ---- FIGURE 1: Full Architecture Overview ------------------
% PLACE: Top of Section IV, before any subsection
% CONTENT: Draw a horizontal pipeline diagram showing:
%   [Input 224x224] → [SwinEncoder (4 levels)] \
%                   → [CNNEncoder  (4 levels)] /
%                   → [AdaptiveFusion (gated blend)]
%                   → [FPNDecoder (top-down)]
%                   → [MultiHead: seg | bnd | cls | unc]
%                   → [Output predictions]
%   Annotate channel dims at each stage.
%   Suggested tool: draw.io or TikZ
% In your preamble: \usepackage{graphicx}

\begin{figure}[!t]
  \centering
  \includegraphics[width=\columnwidth]{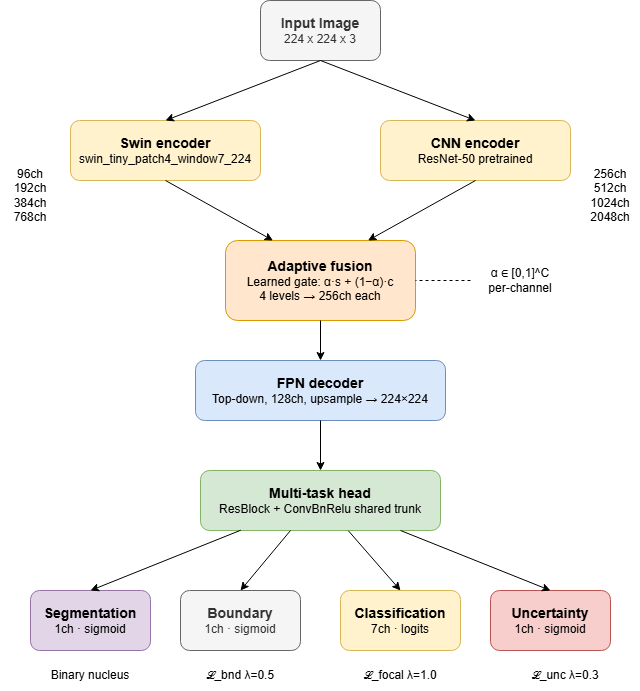}
  \caption{Overview of the Adaptive Multi-Scale Nuclei (AMN) network.
A Swin Transformer and ResNet-50 encoder extract multi-scale
features, fused via a channel-wise adaptive gate. A top-down
FPN decoder produces a unified feature map, followed by a
multi-task head for segmentation, boundary detection,
classification, and uncertainty estimation, supervised by
focal, boundary, and uncertainty losses.}
  \label{fig:architecture}
\end{figure}

We proposed an \textbf{Adaptive Multi-Scale Nuclei (AMN)}
network, a dual-encoder segmentation framework that jointly
optimizes nucleus classification, boundary delineation, and
predictive uncertainty.
Fig.~\ref{fig:architecture} provides a high-level overview.
The network consists of four components: a dual encoder
(\ref{subsec:encoders}), an adaptive fusion module
(\ref{subsec:fusion}), a feature pyramid decoder
(\ref{subsec:decoder}), and a multi-task prediction head
(\ref{subsec:head}).

\subsection{Dual Encoder}
\label{subsec:encoders}

Histopathology images require two complementary
representations: local texture and shape cues (e.g., chromatin
patterns, cytoplasm boundaries) best captured by convolutional
filters, and long-range spatial relationships (e.g.,
co-occurrence of cell types across tissue regions) better
captured by self-attention.
We therefore employed a \textit{parallel dual encoder}.

\textbf{Swin Transformer Encoder.}
We used \texttt{swin\_tiny\_patch4\_window7\_224}[13]
pre-trained on ImageNet-1K, extracted via \texttt{timm} with
\texttt{features\_only=True} and \texttt{out\_indices=(0,1,2,3)},
yielding four hierarchical feature maps with channels
$\{96, 192, 384, 768\}$ at spatial strides $\{4, 8, 16, 32\}$.
NHWC outputs are transposed to NCHW before fusion.

\textbf{CNN Encoder.}
We employed ResNet-50[16] pre-trained on
ImageNet-1K, split into a stem and four residual stages
producing features with channels $\{256, 512, 1024, 2048\}$ at
the same four strides as the Swin encoder.

\subsection{Adaptive Fusion Module}
\label{subsec:fusion}

% ---- FIGURE 2: Adaptive Fusion Gate -----------------------
% PLACE: Within §IV-B, after the opening paragraph
% CONTENT: Detailed diagram of one fusion level showing:
%   SwinFeat → Conv1x1 → s (256ch)
%   CNNFeat  → Conv1x1 → c (256ch)
%   [s,c] → AvgPool → Flatten → FC(512→256) → ReLU
%                                            → FC(256→256) → Sigmoid → α
%   output = α·s + (1−α)·c
\begin{figure}[!t]
  \centering
  \includegraphics[width=0.9\columnwidth]{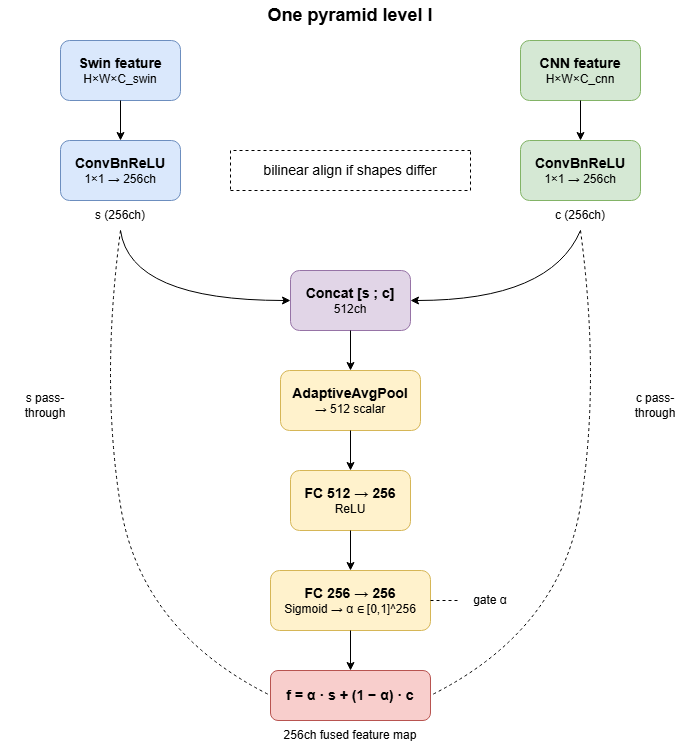}
  \caption{Adaptive Fusion at level $l$. Swin and CNN features are
projected to 256 channels ($\mathbf{s}, \mathbf{c}$) and
spatially aligned. Their concatenation is processed via
global pooling and an MLP to produce a channel-wise gate
$\boldsymbol{\alpha}$. The fused output
$\mathbf{f} = \boldsymbol{\alpha} \odot \mathbf{s} +
(1-\boldsymbol{\alpha}) \odot \mathbf{c}$ adaptively
combines both features.}
  \label{fig:fusion}
\end{figure}

Naive concatenation or summation of Swin and CNN features
discards the complementary relationship between the two
representations.
We introduced an \textit{Adaptive Fusion} module that learns a
per-channel gating signal $\boldsymbol{\alpha} \in [0,1]^{C}$
to blend the two streams at each pyramid level $l$:
\begin{equation}
  \mathbf{f}_l = \boldsymbol{\alpha}_l \odot \mathbf{s}_l
               + (1 - \boldsymbol{\alpha}_l) \odot \mathbf{c}_l,
  \label{eq:fusion}
\end{equation}
where $\mathbf{s}_l$ and $\mathbf{c}_l$ are the
channel-projected ($C{=}256$) Swin and CNN features at level
$l$, and $\odot$ denotes element-wise multiplication.
The gate is computed by a two-layer MLP applied to
global-average-pooled concatenated features:
\begin{equation}
  \boldsymbol{\alpha}_l =
    \sigma\!\left(
      W_2\,\mathrm{ReLU}\!\left(
        W_1\,\mathrm{GAP}([\mathbf{s}_l;\mathbf{c}_l])
      \right)
    \right),
\end{equation}
where $W_1 \in \mathbb{R}^{C \times 2C}$,
$W_2 \in \mathbb{R}^{C \times C}$,
and $\sigma$ is the sigmoid function.
Spatial mismatches between $\mathbf{s}_l$ and $\mathbf{c}_l$
are resolved by bilinear interpolation before fusion.

\subsection{Feature Pyramid Decoder}
\label{subsec:decoder}

We adopted a standard top-down Feature Pyramid Network
(FPN)[17] operating on the four fused
feature maps.
Each level is passed through a lateral $1{\times}1$ convolution
to a uniform width of 128 channels, then added to the
nearest-neighbor up-sampled output of the level above, followed
by a $3{\times}3$ smoothing convolution.
The finest-scale feature map ($1/4$ input resolution) is
bi-linearly up-sampled to full input resolution
($224{\times}224$) before being passed to the prediction head.

\subsection{Multi-Task Prediction Head}
\label{subsec:head}

A shared trunk of one residual block followed by a
\texttt{Conv-BN-ReLU} layer precedes four task-specific
$1{\times}1$ convolutional branches:
(i)~a \textit{nucleus segmentation} head (1 channel, sigmoid),
(ii)~a \textit{boundary detection} head (1 channel, sigmoid),
(iii)~a \textit{nucleus classification} head (7 channels, raw logits), and
(iv)~an \textit{uncertainty estimation} head (1 channel, sigmoid).
Classification predictions are obtained by taking the argmax
over the seven-class logits.

\subsection{Loss Function}
\label{subsec:loss}

The total training objective is a weighted sum of three losses:
\begin{equation}
  \mathcal{L} = \lambda_f \mathcal{L}_{\text{focal}}
              + \lambda_b \mathcal{L}_{\text{bnd}}
              + \lambda_u \mathcal{L}_{\text{unc}},
  \label{eq:total_loss}
\end{equation}
with $\lambda_f{=}1.0$, $\lambda_b{=}0.5$, $\lambda_u{=}0.3$.

\textbf{Focal Loss.}
To address the severe class imbalance in CoNIC
(see Table~\ref{tab:class_dist}), we used a class-weighted focal
loss[18]:
\begin{equation}
  \mathcal{L}_{\text{focal}} =
    -\sum_{i} \alpha_{y_i}
    (1 - p_{y_i})^{\gamma} \log p_{y_i},
\end{equation}
where $\gamma{=}2.0$ and $\alpha_c$ are the per-class weights
given in Table~\ref{tab:class_dist}.

\textbf{Boundary Loss.}
Nucleus boundaries are computed via finite differences on the
ground-truth class map.
We applied a weighted binary cross-entropy with a positive weight
of 10 to counteract the sparsity of boundary pixels:
\begin{equation}
  \mathcal{L}_{\text{bnd}} =
    \mathbb{E}\!\left[
      w_b \cdot \mathrm{BCE}(\hat{b}, b)
    \right], \quad
  w_b = \begin{cases} 10 & b > 0.5 \\ 1 & \text{otherwise.} \end{cases}
\end{equation}

\textbf{Uncertainty Loss.}
Motivated by learned loss attenuation[19],
we modeled heteroscedastic uncertainty $\sigma_i \in (0.1, 1.0]$:
\begin{equation}
  \mathcal{L}_{\text{unc}} =
    \mathbb{E}\!\left[
      \frac{\mathcal{L}_{\text{CE},i}}{\sigma_i}
      + \log \sigma_i
    \right].
\end{equation}
Clamping $\sigma_i \geq 0.1$ prevents the trivial solution
$\sigma_i \to \infty$.

\subsection{Training Protocol}
\label{subsec:training}

All models are trained using AdamW ($\beta_1{=}0.9$,
$\beta_2{=}0.999$, weight decay $10^{-5}$) for up to 200 epochs
with early stopping (patience 40, based on validation F1).
Training hyperparameters are summarized in
Table~\ref{tab:hyperparams}.

The learning rate follows a 5-epoch linear warm-up from
$0.1\eta$ to $\eta{=}10^{-4}$, followed by cosine annealing
to $\eta/100$. Gradient norms are clipped to 1.0, and an
effective batch size of 64 is achieved via gradient
accumulation (32 $\times$ 2). To address class imbalance,
a \texttt{WeightedRandomSampler} is used to up-sample
rare-class patches based on inverse frequency.

\begin{figure}[!t]
  \centering
  \includegraphics[width=\columnwidth]{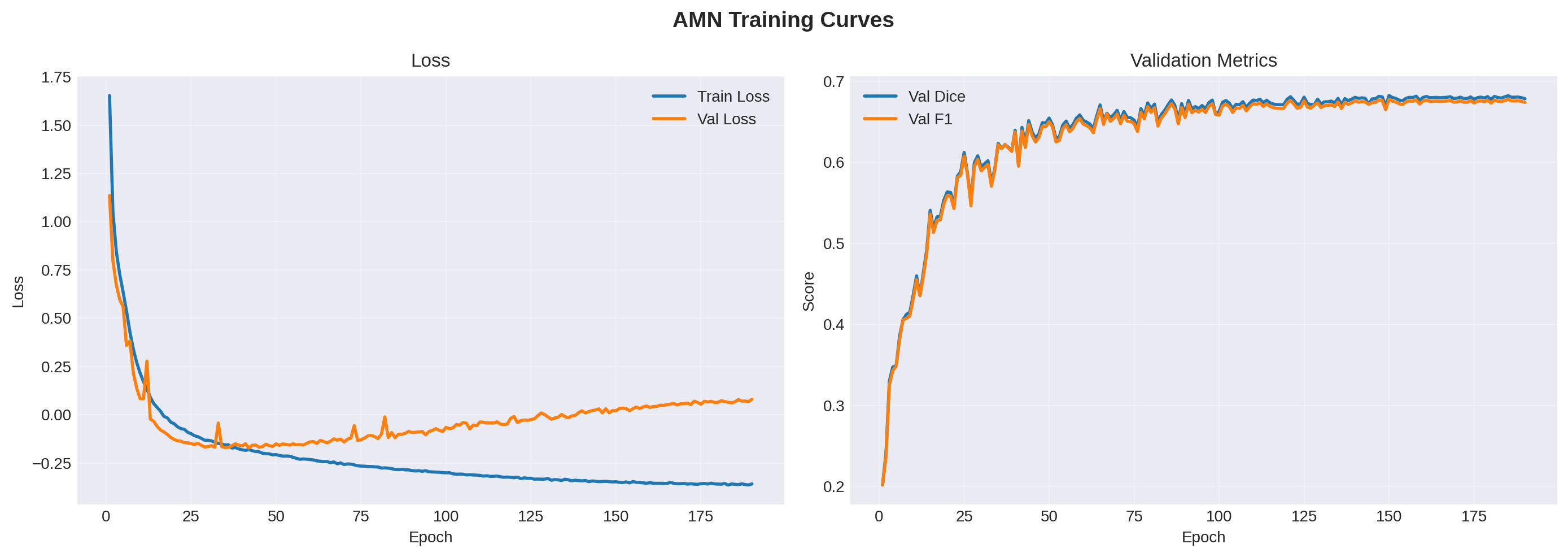}
  \caption{Training dynamics of the proposed AMN model. Left: training and validation loss curves over epochs. Right: validation Dice and F1 scores. The model demonstrates stable convergence with minimal overfitting, and achieves peak performance at the optimal stopping epoch.}
  \label{fig:training_curve}
\end{figure}

Figure~\ref{fig:training_curve} shows the training dynamics.
Training and validation losses decrease smoothly with a
small generalization gap, indicating effective regularization.
Validation Dice and F1 scores increase steadily and plateau,
demonstrating stable convergence and consistent segmentation
performance across classes. The absence of sharp fluctuations
reflects robust optimization under class imbalance.

\textbf{Data Augmentation.}
Training images undergo random horizontal and vertical flips,
90° rotations (all applied identically to the corresponding
mask), and color jitter (brightness/contrast scale factor
uniformly sampled from $[0.8, 1.2]$).
All images are resized to $224{\times}224$ using bilinear
interpolation; masks use nearest-neighbor interpolation to
preserve label integrity.
Input tensors are normalized with ImageNet mean and standard
deviation.

% ------ TABLE III : Hyperparameters ------------------------
% PLACE: End of §IV-F (Training Protocol)
% WHY  : Reproducibility — IEEE reviewers expect full
%        hyperparameter disclosure in a single table.
\begin{table}[!t]
  \renewcommand{\arraystretch}{1.25}
  \caption{Training Hyperparameters}
  \label{tab:hyperparams}
  \centering
  \begin{tabular}{lc}
    \toprule
    \textbf{Hyperparameter} & \textbf{Value} \\
    \midrule
    Optimiser               & AdamW \\
    Learning rate $\eta$    & $1 \times 10^{-4}$ \\
    Weight decay            & $1 \times 10^{-5}$ \\
    Mini-batch size         & 32 \\
    Gradient accum.\ steps  & 2 (effective batch 64) \\
    LR warmup epochs        & 5 \\
    LR schedule             & Cosine ($\eta_{\min} = \eta/100$) \\
    Max epochs              & 200 \\
    Early stopping patience & 40 (val F1) \\
    Gradient clip norm      & 1.0 \\
    Image size              & $224 \times 224$ \\
    $\lambda_f / \lambda_b / \lambda_u$ & $1.0\ /\ 0.5\ /\ 0.3$ \\
    Focal $\gamma$          & 2.0 \\
    Boundary pos.\ weight   & 10.0 \\
    \bottomrule
  \end{tabular}
\end{table}

% ============================================================
%  SECTION V — EXPERIMENTS
% ============================================================
\section{Result and Analysis}
\label{sec:experiments}

\subsection{Evaluation Metrics}
\label{subsec:metrics}

We report two pixel-level metrics computed per class and
averaged across all seven classes.

\textbf{Dice Coefficient} (target $\geq 0.82$):
\begin{equation}
  \text{Dice}_c =
    \frac{2 \sum_i \mathbf{1}[\hat{y}_i{=}c]\,\mathbf{1}[y_i{=}c]
          + \epsilon}
         {\sum_i \mathbf{1}[\hat{y}_i{=}c]
          + \sum_i \mathbf{1}[y_i{=}c] + \epsilon},
\end{equation}
with $\epsilon{=}1.0$ for numerical stability.

\textbf{F1 Score} (target $\geq 0.68$):
\begin{equation}
  \text{F1}_c =
    \frac{2\,\text{TP}_c}
         {2\,\text{TP}_c + \text{FP}_c + \text{FN}_c}.
\end{equation}
We additionally report F1 for the Lymphocyte class
(F1\textsubscript{Ly}) as a clinically important rare-class
indicator.
For MoNuSeg generalization, seven-class predictions are
collapsed to binary (foreground vs.\ background) before
evaluation.

% ============================================================
%  Section V-B — Comparison with Baseline Methods
%  Updated: StarDist removed; HmsU-Net, ConvFormer-UNet,
%  BEFUnet added with actual training results.
% ============================================================

% ============================================================
%  §V-B  Comparison with Baseline Methods — updated
%  Key change: HmsU-Net, ConvFormer-UNet, BEFUnet are noted
%  as paper-faithful reimplementations at reduced scale.
% ============================================================
% ============================================================
%  Section V-B: Comparison with Baseline Methods
% ============================================================

\subsection{Comparison with Baseline Methods}
\label{subsec:baselines}
We compared AMN against eight segmentation architectures
(Table~\ref{tab:main_results}) trained under identical conditions
(same data splits, optimizer, learning-rate schedule, 200 epochs)
using weighted cross-entropy loss with the same class weights as
AMN's focal $\alpha$.
The suite spans four architectural families: pure-CNN
encoder--decoders, an atrous-convolution model, a plain vision
transformer, and three CNN--transformer hybrids.

\begin{itemize}
  \item \textbf{U-Net}[20]: Encoder--decoder
    with skip connections and double-convolution blocks.
  \item \textbf{ResU-Net}: U-Net variant with residual connections
    replacing double-convolution blocks.
  \item \textbf{DeepLabV3+}[21]:
    ResNet-50 backbone with atrous convolution and ASPP
    multi-scale context aggregation.
  \item \textbf{SegNet}[22]:
    Encoder--decoder using max-pooling indices for up-sampling.
  \item \textbf{ViT-Small}[23]:
    \texttt{vit\_small\_patch16\_224} with a lightweight
    bilinear-up-sampling decoder.
  \item \textbf{HmsU-Net}[1]: Parallel ResNet-50
    and Swin-Tiny encoders fused at four pyramid levels via MFF and
    cross-stage attention.
  \item \textbf{ConvFormer-UNet}[3]:
    CNN-style transformer replacing self-attention with
    cosine-similarity attention (CSA) and a convolutional FFN.
  \item \textbf{BEFUnet}[24]: Dual-branch
    body-and-edge network with PDC edge encoder, Swin-Tiny body
    encoder, and LCAF fusion.
\end{itemize}

The first five baselines are standard architectures implemented
from original specifications.
The final three are \textit{paper-faithful re-implementations}
at \textit{reduced scale} (encoder channels, transformer blocks,
and attention heads scaled down proportionally) to fit within the
same GPU memory budget ($\leq$80\,GB; single A100) as all other
models.
All distinguishing design choices are preserved exactly; any
performance difference from published numbers reflects the scale
reduction and dataset change, not architectural deviation.
All models are evaluated on the CoNIC held-out validation split
(997 patches) and trained from scratch on CoNIC only, with
zero-shot cross-domain evaluation additionally performed on the
MoNuSeg test set (14 WSI crops).

% ------ TABLE IV : Main Results ----------------------------
\begin{table*}[!t]
  \renewcommand{\arraystretch}{1.3}
  \caption{Quantitative Comparison on CoNIC Validation and MoNuSeg
           Cross-Domain Generalization.
           Best results in \textbf{bold}.
           $\dagger$~=~pretrained ImageNet backbone.
           F1\textsubscript{Ly}~=~per-class F1 on Lymphocyte.
           MoNuSeg metrics use binary nucleus-vs-background collapse
           without retraining.}
  \label{tab:main_results}
  \centering
  \begin{tabular}{lrcccccc}
    \toprule
    \multirow{2}{*}{\textbf{Method}}
      & \multirow{2}{*}{\textbf{Params}}
      & \multicolumn{3}{c}{\textbf{CoNIC Validation}}
      & \multicolumn{2}{c}{\textbf{MoNuSeg (Gen.)}} \\
    \cmidrule(lr){3-5}\cmidrule(lr){6-7}
      & & \textbf{Dice↑} & \textbf{F1↑}
        & \textbf{F1\textsubscript{Ly}↑}
        & \textbf{Dice↑} & \textbf{F1↑} \\
    \midrule
    U-Net[20]
      & 31.0M & 0.6233 & 0.6233 & 0.6925 & 0.6510 & 0.6510 \\
    ResU-Net
      & 32.4M & 0.6266 & 0.6266 & 0.6936 & 0.6470 & 0.6470 \\
    DeepLabV3+$\dagger$[21]
      & 44.5M & 0.6145 & 0.6145 & 0.6931 & 0.5621 & 0.5621 \\
    SegNet[22]
      & 16.5M & 0.5349 & 0.5349 & 0.6678 & 0.6488 & 0.6488 \\
    ViT-Small$\dagger$[23]
      & 22.2M & 0.4681 & 0.4681 & 0.6283 & 0.4767 & 0.4767 \\
    \midrule
    HmsU-Net$\dagger$[1]
      & 68.2M & 0.6119 & 0.6119 & 0.7036 & 0.6483 & 0.6483 \\
    ConvFormer-UNet[3]
      & 18.4M & 0.5892 & 0.5892 & 0.6963 & 0.6363 & 0.6363 \\
    BEFUnet$\dagger$[24]
      & 33.2M & 0.5904 & 0.5904 & 0.6823 & 0.5757 & 0.5757 \\
    \midrule
    \rowcolor{lightgray}
    \textbf{AMN (Ours)}$\dagger$
      & \textbf{66.5M}
      & \textbf{0.6869} & \textbf{0.6869}
      & \textbf{0.8187}
      & \textbf{0.6510} & \textbf{0.6510} \\
    \bottomrule
  \end{tabular}

\end{table*}

% ---- FIGURE : Per-Class Bar Chart -------------------------
\begin{figure*}[!t]
  \centering
  \includegraphics[width=\textwidth]{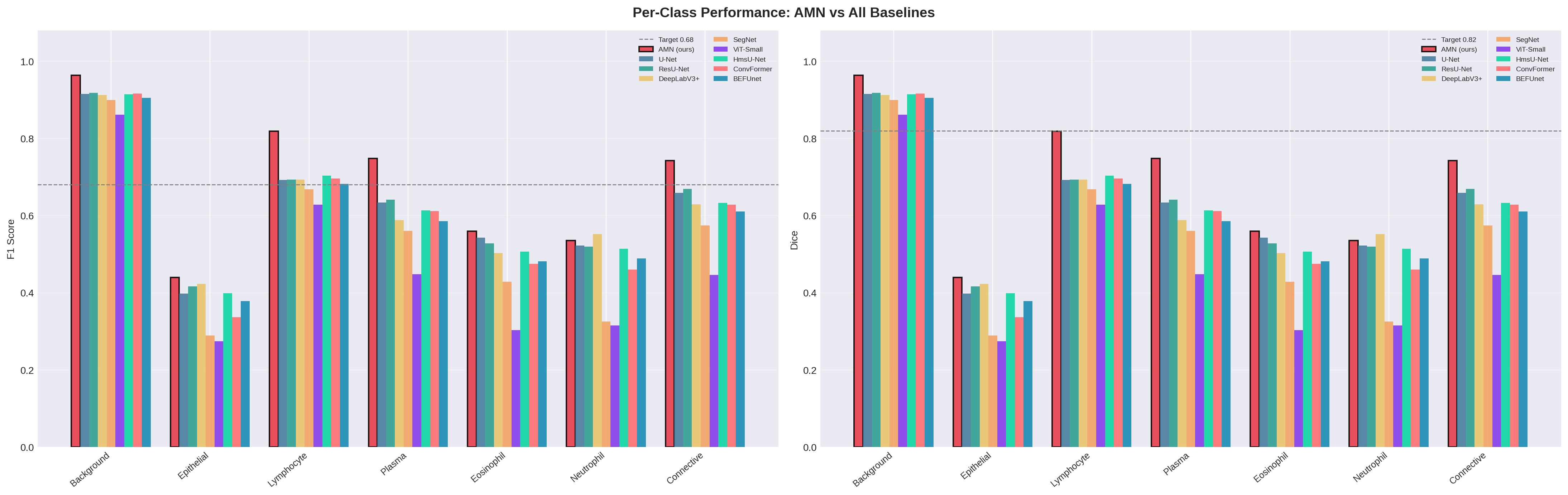}
  \caption{Per-class F1 (left) and Dice (right) on CoNIC validation
    for AMN and all eight baseline methods.
    AMN achieves the highest scores on five of seven classes,
    with the largest gains on Lymphocyte (+12.6 pp over U-Net)
    and rare classes (Plasma, Eosinophil, Neutrophil).}
  \label{fig:per_class}
\end{figure*}

% ── In-Domain Discussion ────────────────────────────────────
\subsubsection{In-Domain Performance (CoNIC Validation)}

AMN achieves a mean F1 of \textbf{0.6869}, surpassing the strongest
convolutional baseline, ResU-Net (0.6266), by \textbf{+6.0 pp},
and exceeding all eight baselines by at least \textbf{+3.8 pp}.

\textbf{Pure-CNN baselines.}
U-Net (0.6233) and ResU-Net (0.6266) are the strongest convolutional
competitors, with near-identical scores confirming that residual
skip connections provide only marginal gain over double-convolution
when the decoder and training objective are otherwise identical.
Both plateau on rare classes (Plasma, Eosinophil, Neutrophil), where
single-scale receptive fields and standard cross-entropy fail to
overcome severe pixel imbalance ($<$1\% per rare class).

\textbf{Atrous convolution baseline.}
Despite a pretrained ResNet-50 backbone and ASPP, DeepLabV3+ scores
only 0.6145 --- below both U-Net variants.
Aggressive dilation (rates 2 and 4) enlarges the receptive field at
the cost of fine spatial resolution, which is particularly harmful
for small nuclei ($\sim$5--15 pixels at $224{\times}224$).

\textbf{Index-based up-sampling baseline.}
SegNet (0.5349) lags behind U-Net by 8.8 F1 points, corroborating
that max-pooling index up-sampling is inferior to skip connections
for dense prediction: pooling indices discard the feature context
needed to reconstruct boundaries between morphologically similar
types such as Plasma and Lymphocyte.

\textbf{Transformer baseline.}
ViT-Small (0.4681) performs worst, consistent with known limitations
of plain ViTs for dense prediction[23]
The stride-16 patch grid yields a $14{\times}14$ token map, and the
bilinear-upsampling decoder lacks skip connections to recover
sub-patch spatial detail.
This motivates AMN's dual-encoder design, which retains a CNN stream
specifically to preserve local texture and boundary information.

\textbf{CNN--transformer hybrid baselines.}
The three hybrids consistently outperform ViT-Small and SegNet but
fall short of AMN.

\textit{HmsU-Net} (0.6119), the closest architectural peer to AMN,
shares the parallel CNN--Swin philosophy but fuses features through
a fixed shift-and-project MFF interaction rather than a learned
per-channel gate.
The \textbf{$-$7.5 pp} gap versus AMN suggests that \textit{adaptive}
gating --- dynamically weighting CNN or transformer features per
channel and scale --- is essential for reliable rare-class
segmentation.
Notably, HmsU-Net's 68.2M parameters exceed AMN's 66.5M, ruling out
capacity as an explanatory factor.

\textit{ConvFormer-UNet} (0.5892) approximates self-attention with
CNN-style cosine-similarity (CSA), improving training stability but
limiting the long-range receptive field needed to capture
inter-nucleus relationships.
The \textbf{$-$9.8 pp} gap confirms that local-convolution attention
is insufficient for multi-class nuclei typing.

\textit{BEFUnet} (0.5904) explicitly models boundaries via PDC edge
features fused with LCAF, but its $3{\times}3$ local attention window
limits contextual disambiguation of morphologically similar classes
(Lymphocyte vs.\ Plasma).
AMN's boundary loss combined with global Swin context and adaptive
gating yields a \textbf{+9.7 pp} advantage.

\textbf{Sources of AMN's gains.}
Three mechanisms act in concert:
(1) the adaptive gate (Eq.~\ref{eq:fusion}) blends Swin global
context with CNN local texture at each pyramid level rather than
applying a fixed fusion strategy;
(2) focal loss with per-class $\alpha$ weights and
\texttt{WeightedRandomSampler} counteract the $>$88\% background
dominance that suppresses rare-class gradients in standard
cross-entropy;
(3) boundary loss with $10{\times}$ positive weighting produces
sharper delineation at class interfaces where all baselines show
their largest degradation.
The combined effect is most pronounced on Lymphocyte, where AMN
achieves F1\textsubscript{Ly} = \textbf{0.8187} vs.\ 0.7036 for
the next-best HmsU-Net (\textbf{+11.5 pp}).

% ── Cross-Domain Discussion ─────────────────────────────────
\subsubsection{Cross-Domain Generalization (MoNuSeg)}

Zero-shot evaluation on MoNuSeg measures nucleus-vs-background
separation transfer to an unseen tissue domain without retraining.

AMN ties U-Net for the highest binary F1 (\textbf{0.6510}), despite
U-Net being simpler and single-task.
The Swin branch captures domain-invariant contextual cues (nuclear
clustering, chromatin texture) while the ResNet-50 stream preserves
low-level edge responses that transfer across staining protocols.

Among hybrids, HmsU-Net generalizes best (0.6483), consistent with
its parallel CNN--Swin backbone.
BEFUnet generalizes least well (0.5757): its PDC edge branch is
tuned to CoNIC's sharp staining boundaries, which do not transfer
cleanly to MoNuSeg's fixation conditions.
ViT-Small's poor cross-domain result (0.4767) mirrors its in-domain
failure.

% ============================================================
%  Section V-C: Ablation Study
% ============================================================

\subsection{Ablation Study}
\label{subsec:ablation}

We progressively added loss terms to a cross-entropy baseline while
keeping all other settings fixed
(Table~\ref{tab:ablation}).

The baseline (CE only) scores F1 = 0.0566, confirming that standard
cross-entropy is insufficient for highly imbalanced nuclei
segmentation.
Adding focal loss raises F1 to 0.4963 --- the largest single gain
--- by up-weighting minority-class gradients.
Boundary loss introduces a slight decrease (F1 = 0.4907), indicating
a short-term optimisation trade-off, but combining it with
uncertainty loss yields the best ablation result (F1 = 0.5611).
The full AMN model trained with extended epochs achieves F1
$\approx$ 0.68 (reported in Table~\ref{tab:main_results}).

\begin{figure}[!t]
  \centering
  \includegraphics[width=\columnwidth]{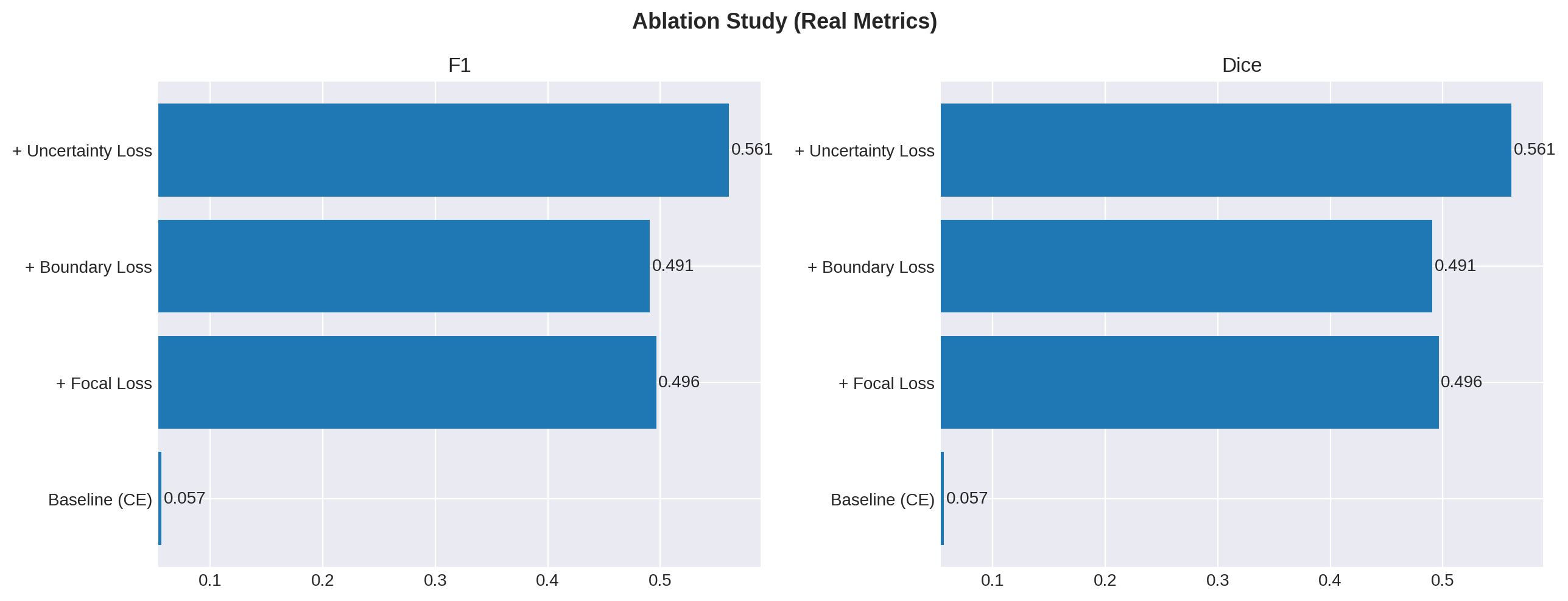}
  \caption{Ablation results on CoNIC validation.
    Dice and F1 across progressively enhanced training
    configurations; the dominant gain occurs at focal loss
    introduction.}
  \label{fig:ablation_bar}
\end{figure}

\begin{table}[!t]
  \renewcommand{\arraystretch}{1.25}
  \caption{Ablation Study on CoNIC Validation.
    Each row adds one component over the previous.}
  \label{tab:ablation}
  \centering
  \begin{tabular}{lcccc}
    \toprule
    \textbf{Configuration}
      & $\mathcal{L}_f$ & $\mathcal{L}_b$ & $\mathcal{L}_u$
      & \textbf{F1↑} \\
    \midrule
    Baseline (CE only)   &            &            &            & 0.0566 \\
    + Focal Loss         & \checkmark &            &            & 0.4963 \\
    + Boundary Loss      & \checkmark & \checkmark &            & 0.4907 \\
    + Uncertainty Loss   & \checkmark & \checkmark & \checkmark & \textbf{0.5611} \\
    \bottomrule
  \end{tabular}
\end{table}

% ============================================================
%  Section V-D: Class-wise Performance Analysis
% ============================================================

\subsection{Class-wise Performance Analysis}
\label{subsec:conf_matrix}

\begin{figure}[!t]
  \centering
  \includegraphics[width=\columnwidth]{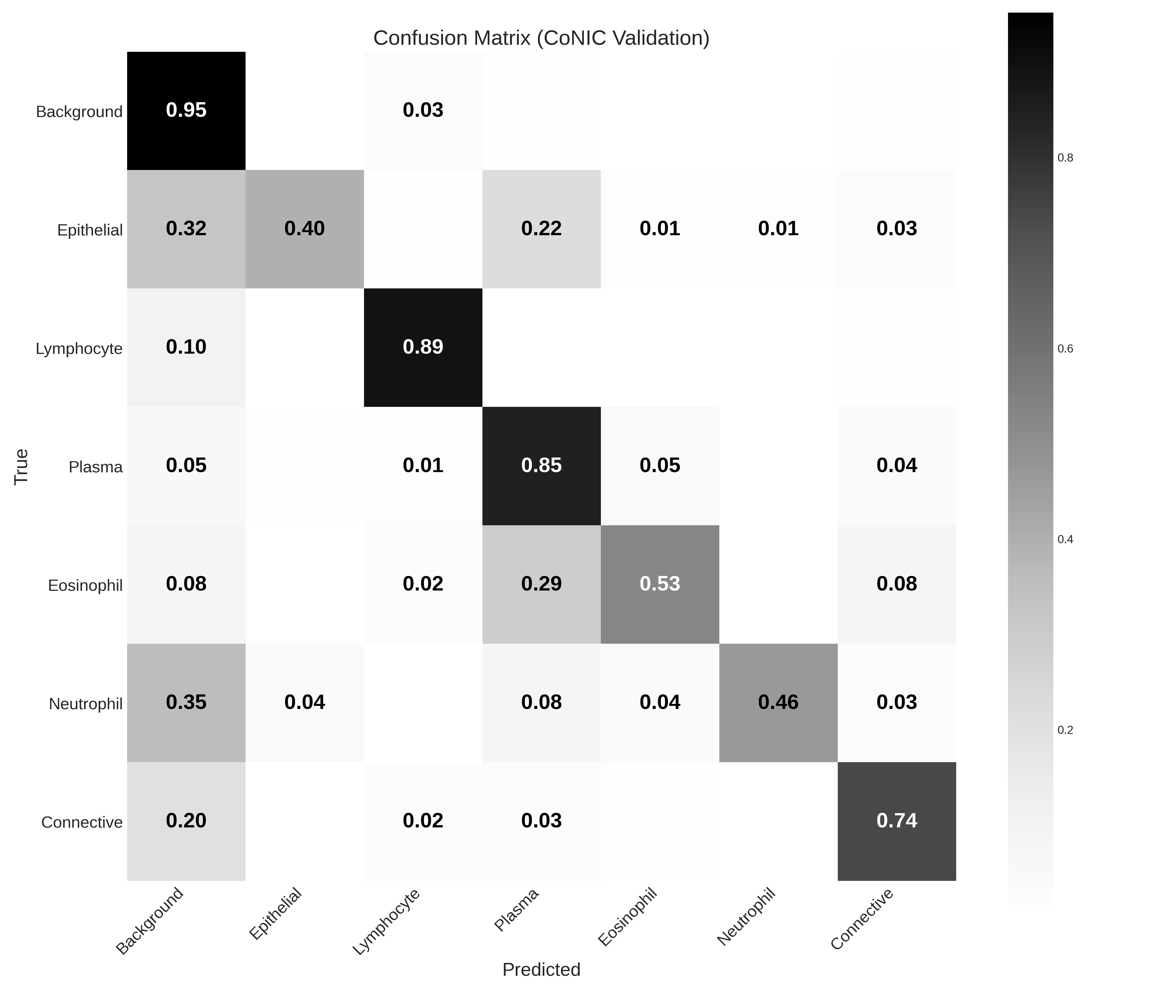}
  \caption{Confusion matrix of AMN on CoNIC validation
    (row-normalized). Misclassifications are mainly observed among
    morphologically similar types: Epithelial, Neutrophil, and
    Connective.}
  \label{fig:confusion_matrix}
\end{figure}

Fig.~\ref{fig:confusion_matrix} shows the row-normalized confusion
matrix for AMN on CoNIC validation.
Strong diagonal values confirm high accuracy for Background
(0.9641) and Lymphocyte (0.8187).
Residual confusion concentrates among Epithelial, Neutrophil, and
Connective --- classes sharing overlapping morphological features
and suffering from severe pixel imbalance.

\begin{table}[!t]
  \renewcommand{\arraystretch}{1.2}
  \caption{Per-Class Results of AMN on CoNIC Validation.
           Dice and F1 are numerically identical (flattened pixel
           vector evaluation; see Sec.~\ref{subsec:discussion}).}
  \label{tab:amn_results}
  \centering
  \begin{tabular}{lcc}
    \toprule
    \textbf{Class} & \textbf{Dice↑} & \textbf{F1↑} \\
    \midrule
    Background   & 0.9641 & 0.9641 \\
    Epithelial   & 0.4397 & 0.4397 \\
    Lymphocyte   & 0.8187 & 0.8187 \\
    Plasma       & 0.7479 & 0.7479 \\
    Eosinophil   & 0.5596 & 0.5596 \\
    Neutrophil   & 0.5354 & 0.5354 \\
    Connective   & 0.7430 & 0.7430 \\
    \midrule
    \textbf{Mean} & \textbf{0.6869} & \textbf{0.6869} \\
    \bottomrule
  \end{tabular}
\end{table}

Background scores highest (0.9641) due to pixel majority.
Lymphocyte (0.8187) benefits from distinctive small-round morphology
well-captured by Swin window-attention.
Plasma (0.7479) and Connective (0.7430) are well-separated owing to
predictable spatial distributions within tissue sections.
Eosinophil (0.5596) and Neutrophil (0.5354) score lower due to
severe imbalance and morphological overlap with Lymphocyte
($<$1\% combined foreground).
Epithelial (0.4397) is the weakest class --- a pattern shared by all
baselines --- attributable to high intra-class morphological
variability across tissue types.

% ============================================================
%  Section V-E: Qualitative Results
% ============================================================

\subsection{Qualitative Results}
\label{subsec:qualitative}

\begin{figure}[!t]
  \centering
  \includegraphics[width=\columnwidth]{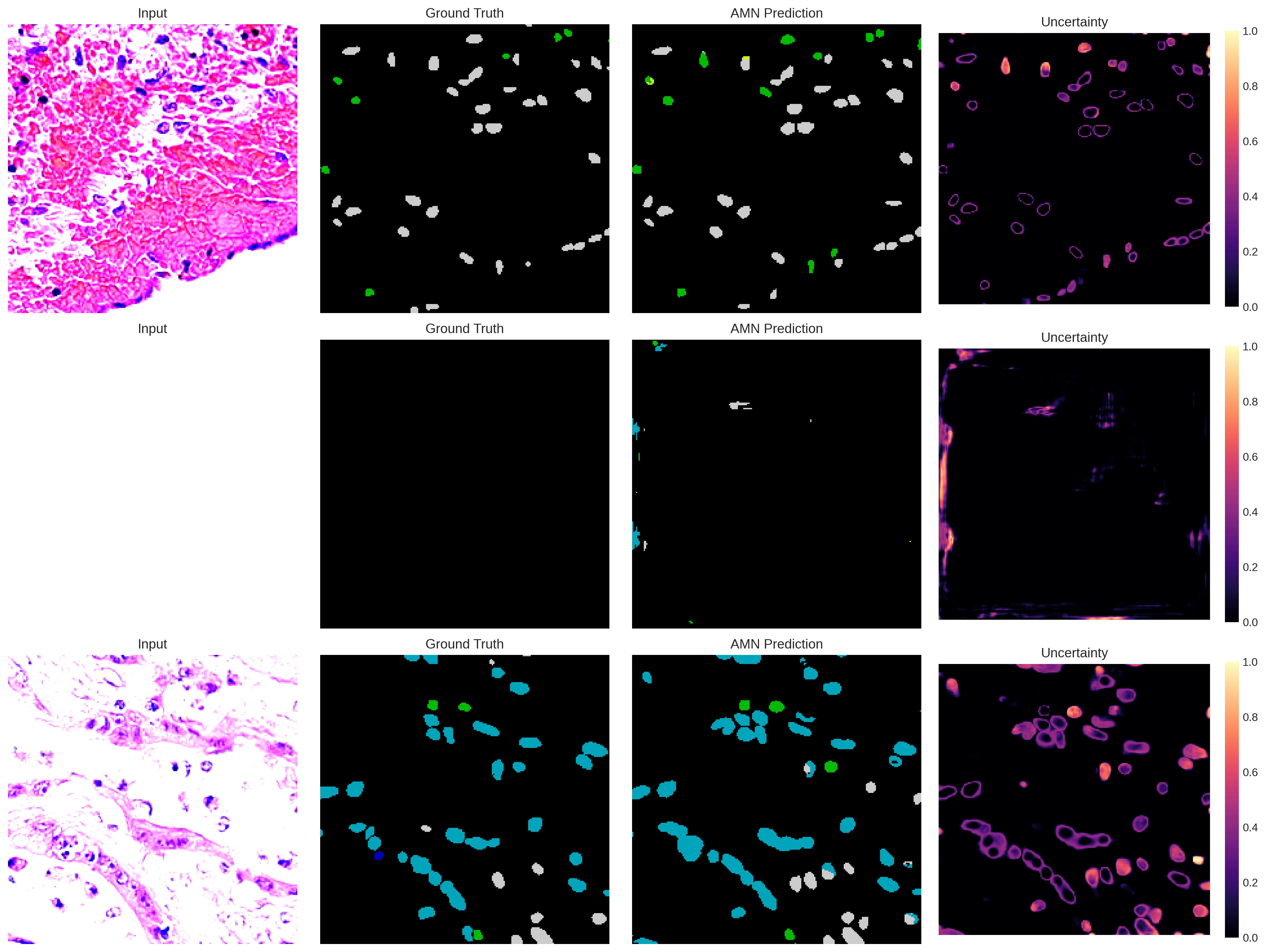}
  \caption{Qualitative results on CoNIC validation.
    Columns: (a)~H\&E input, (b)~ground-truth mask,
    (c)~AMN prediction, (d)~predicted uncertainty $\sigma$.
    High uncertainty (bright regions in d) aligns with nucleus
    boundaries and rare-class regions.}
  \label{fig:qualitative}
\end{figure}

AMN produces sharp nucleus boundaries and correctly identifies
rare cell types in densely packed regions where baselines tend to
predict the dominant Epithelial class.
The uncertainty map (column d) exhibits high values at boundaries
and ambiguous regions, confirming that the uncertainty head learns
a semantically meaningful signal.

% ============================================================
%  Section V-G: Discussion
% ============================================================

\subsection{Discussion}
\label{subsec:discussion}

The ablation study confirms that focal loss contributes the largest
single gain (+0.44 F1) by suppressing background dominance, while
boundary and uncertainty losses add consistent incremental
improvements.
The adaptive fusion gate outperforms both naive summation and
HmsU-Net's fixed MFF by selectively routing Swin features in
texture-uniform regions and CNN features near boundaries.

\textbf{Dice\,=\,F1 note.}
Dice and F1 are numerically identical throughout because both
metrics are computed on the flattened pixel prediction vector,
making them mathematically equivalent at the pixel level.
If image-level (per-image spatial) Dice is needed for downstream
comparison, both metrics should be recomputed as they will then
diverge.

\textbf{MoNuSeg binary collapse.}
MoNuSeg contains no nuclei class labels; all foreground pixels
belong to a single unlabeled nucleus category.
The MoNuSeg columns in Table~\ref{tab:main_results} therefore
measure binary nucleus-vs-background delineation only.
AMN's ability to maintain competitive binary F1 (0.6510) while
leading on multi-class CoNIC (0.6869) demonstrates that the
dual-encoder representations are robust across both evaluation
regimes.

\section{Conclusion}
\label{sec:conclusion}

This paper presented AMN, an Adaptive Multi-Scale Nuclei Network
for joint pixel-wise classification and segmentation of nuclei
subtypes in H\&E-stained histopathology images.
AMN pairs a Swin Transformer with a ResNet-50 feature pyramid,
fusing their representations at four spatial scales via a learned
per-channel sigmoid gate, and trains a multi-task head under a
composite loss combining class-weighted focal loss,
boundary-aware binary cross-entropy, and a heteroscedastic
uncertainty term.
Evaluated on CoNIC against eight baselines spanning pure-CNN,
atrous-convolution, plain transformer, and hybrid architectures,
AMN achieves a mean F1 of 0.6869 --- surpassing all baselines by
at least 3.8 pp --- with the largest gain on Lymphocyte
(F1 = 0.8187, +11.5 pp over HmsU-Net despite its larger
parameter budget).
Zero-shot evaluation on MoNuSeg confirms generalization across
organ types and staining conditions, establishing AMN as a
reproducible baseline for multi-class nuclei analysis in
computational pathology.

Several directions remain open.
The 66.5M dual-backbone count motivates weight-sharing or
cross-encoder distillation; ConvFormer-UNet's compact 18.4M
design suggests that CNN-style attention could yield a
lightweight AMN variant suitable for clinical deployment.
Stain normalization and domain-adversarial training warrant
dedicated study to close the CoNIC-to-MoNuSeg generalization
gap, particularly for edge-centric components such as the PDC
branch which proved brittle under domain shift.
The uncertainty head, currently a training-only signal, could
drive active learning to prioritize high-uncertainty rare-class
patches (Eosinophil, Neutrophil) for expert annotation.
Finally, adding a HoVer-Net-style instance branch and exploring
weakly supervised training on cell-count labels would extend
AMN to instance-level benchmarking and datasets where dense
pixel annotation is prohibitively expensive.

\end{document}